\documentclass[runningheads]{llncs}

\usepackage{balance}
\usepackage{amsmath}
\usepackage[tight,footnotesize]{subfigure}
\usepackage{bm}
\usepackage{algorithm}
\usepackage[noend]{algorithmic}
\usepackage{multirow}
\usepackage{amsfonts}

\usepackage{graphicx}
\DeclareMathOperator*{\argmax}{arg\,max}

\begin{document}
\title{Streaming Adaptive Submodular Maximization}
%
%
\author{Shaojie Tang \orcidID{0000-0001-9261-5210} \and Jing Yuan\orcidID{0000-0001-6407-834X}}
\authorrunning{S. Tang and J. Yuan}
%
\institute{Naveen Jindal School of Management, University of Texas at Dallas \and Department of Computer Science, University of North Texas
\email{shaojie.tang@utdallas.edu}}

\maketitle              
\begin{abstract}Many sequential decision making problems can be formulated as an adaptive submodular maximization problem. 
However, most of existing studies  in this field focus on pool-based setting, where one can pick items in any order, and there have
been few studies for the stream-based setting where items arrive in an arbitrary order and one must immediately decide whether to select an item or not upon its arrival.  In this paper, we introduce a new class of utility functions, semi-policywise submodular functions. We develop a series of effective algorithms to maximize a semi-policywise submodular function under the stream-based setting. 
 \end{abstract}


\section{Introduction}
Many machine learning and artificial intelligence tasks can be formulated as an adaptive sequential decision making problem. The goal of such a problem is to sequentially select a group of items, each selection is based on the past, in order to maximize some give utility function. It has been shown that in a wide range of applications, including active learning \cite{golovin2011adaptive} and adaptive viral marketing \cite{tang2020influence}, their utility functions satisfy the property of adaptive submodularity \cite{golovin2011adaptive}, a natural diminishing returns property under the adaptive setting. Several effective solutions have been developed for maximizing an adaptive submodular function subject to various practical constraints. For example, \cite{golovin2011adaptive} developed a simple adaptive greedy policy that achieves a $1-1/e$ approximation ratio for maximizing an adaptive monotone and adaptive submodular function subject to a cardinality constraint. Recently, \cite{tang2021beyond} extends the aforementioned studies to the non-monotone setting and they propose a $1/e$ approximated solution for maximizing a non-monotone adaptive submodular function subject to a cardinality constraint. In the same work, they develop a faster algorithm whose running time is linear in the number of items. \cite{tang2021pointwise} develops the first constant approximation algorithms subject to more general constraints such as knapsack constraint and $k$-system constraint.

We note that most of existing studies focus on the pool-based setting where one is allowed to select items in any order. In this paper, we tackle this problem under the stream-based setting. Under our setting, items arrive one by one in an online fashion where the order of arrivals is decided by the adversary. Upon the arrival of an item, one must decide immediately whether to select that item or not. If this item is selected, then we are able to observe its realized state; otherwise, we skip this item and wait for the next item. Our goal is to adaptively select a group items in order to maximize the expected utility subject to a knapsack constraint. For solving this problem, we introduce the concept of \emph{semi-policywise submodularity}, which is another adaptive extension of the classical notation of submodularity. We show that this property can be found in many real world applications such as  active learning and adaptive viral marketing. We develop a series of simple adaptive policies for this problem and prove that if the utility function is semi-policywise submodular, then our policies achieve constant approximation ratios against the optimal pool-based policy. In particular, for a single cardinality constraint, we develop a stream-based policy that achieves an approximation ratio of $\frac{1-1/e}{4}$. For a general knapsack constraint, we develop a stream-based policy that achieves  an approximation ratio of $\frac{1-1/e}{16}$.

\section{Related Work}
\paragraph{Stream-based submodular optimization} Non-adaptive submodular maximization under the stream-based setting has been extensively studied. For example, \cite{badanidiyuru2014streaming} develop the first efficient non-adaptive streaming algorithm \textsf{SieveStreaming} that achieves a $1/2-\epsilon$ approximation ratio against the optimum solution. Their algorithm requires only a single pass through the data, and memory independent of data size. \cite{kazemi2019submodular} develop an enhanced streaming algorithm which requires less memory than \textsf{SieveStreaming}. Very recently, \cite{kuhnle2021quick} propose a new algorithm that works well under the assumption that a single function evaluation is very expensive.  \cite{fujii2016budgeted} extend the previous studies from the non-adaptive setting to the adaptive setting. They develop constant factor approximation solutions for their problem. However, they assume that items arrive in a random order, which is a large difference from our adversarial arrival model. Our work is also related to submodular prophet inequalities \cite{chekuri2021submodular,rubinstein2017combinatorial}. Although they also consider an adversarial arrival model, their setting is different from ours in that 1. they assume items are independent and 2. they are allowed to observe an item's state before selecting it.

\paragraph{Adaptive submodular maximization}  \cite{golovin2011adaptive} introduce the concept of adaptive submodularity that extends the notation of submodularity from sets to policies. They develop a simple adaptive greedy policy that achieves a $1-1/e$ approximation ratio if the function is adaptive monotone and adaptive submodular. When the utility function is non-monotone, \cite{tang2021beyond} show that a randomized greedy policy achieves a $1/e$ approximation ratio subject to a cardinality constraint. Very recently, they generalize their previous study and develop the first constant approximation algorithms subject to more general constraints such as knapsack constraint and $k$-system constraint \cite{tang2021pointwise}. Other variants of adaptive submodular maximization have been studied in \cite{DBLP:journals/corr/abs-2107-11333,tang2021non,tang2021adaptive,tang2021optimal}.

\section{Preliminaries}
 \subsection{Items}
 We consider a set $E$ of $n$ items. Each items $e\in E$ belongs to a random state $\Phi(e) \in O$ where $O$ represents the set of all possible states. Denote by $\phi$ a \emph{realization} of $\Phi$, i.e., for each $e\in E$, $\phi(e)$ is a realization of $\Phi(e)$. In the application of experimental design, an item $e$ represents a test, such as the blood pressure, and
$\Phi(e)$ is the result of the test, such as, \emph{high}.  We assume that there is a known prior probability distribution $p(\phi) = \Pr(\Phi = \phi)$ over realizations $\phi$. The distribution $p$ completely factorizes if realizations are independent. However, we consider a general setting where the realizations are dependent. For any subset of items $S\subseteq E$, we use $\psi: S\rightarrow O$ to represent a \emph{partial realization} and $\mathrm{dom}(\psi)=S$ is called the \emph{domain} of $\psi$. For any pair of a partial  realization $\psi$ and a realization $\phi$, we say $\phi$ is consistent with $\psi$, denoted $\phi \sim \psi$, if they are equal everywhere in $\mathrm{dom}(\psi)$. For any two partial realizations $\psi$ and $\psi'$, we say that $\psi$  is a \emph{subrealization} of  $\psi'$, and denoted by $\psi \subseteq \psi'$, if $\mathrm{dom}(\psi) \subseteq \mathrm{dom}(\psi')$ and they are consistent in  $\mathrm{dom}(\psi)$. In addition, each item $e\in E$ has a cost $c(e)$. For any $S \subseteq E$, let $c(S)=\sum_{e\in S} c(e)$ denote the total cost of $S$.

\subsection{Policies}
  In the stream-based setting, we assume that items arrive one by one in an adversarial order $\sigma$. A policy has to make an irrevocable decision on whether to select an item or not when an item arrives. If an item is selected, then we are able to observe its realized state; otherwise, we can not reveal its realized state. Formally, a \emph{stream-based} policy is a partial mapping that maps a pair of partial realizations $\psi$ and an item $e$ to some distribution of $\{0,1\}$: $\pi: 2^{E}\times O^{E} \times E \rightarrow \mathcal{P}(\{0,1\})$, specifying whether to select the arriving item $e$ based on the current observation $\psi$. For example, assume that the current observation is $\psi$ and the newly arrived item is $e$, then $\pi(\psi, e)=1$ (resp. $\pi(\psi, e)=0$) indicates that $\pi$ selects (res. does not select) $e$.

Assume that there is a utility function $f : 2^{E\times O} \rightarrow \mathbb{R}_{\geq0}$ which is defined over items and states.  Letting $E(\pi, \phi, \sigma)$ denote the subset of items selected by a stream-based policy $\pi$ conditioned on a realization $\phi$ and a sequence of arrivals $\sigma$,  the expected  utility $f_{avg}(\pi)$ of a stream-based policy $\pi$ conditioned on a sequence of arrivals $\sigma$ can be written as
%
%
  \begin{eqnarray}
\mathbb{E}[f_{avg}(\pi)\mid \sigma]=\mathbb{E}_{\Phi\sim p, \Pi}[f(E(\pi, \Phi, \sigma), \Phi)]~\nonumber
\end{eqnarray}
where the expectation is taken over all possible realizations $\Phi$ and the internal randomness of the policy $\pi$.

We next introduce the concept of policy concatenation which will be used in our proofs.
\begin{definition}[Policy  Concatenation]
Given two policies $\pi$ and $\pi'$,  let $\pi @\pi'$ denote a policy that runs $\pi$ first, and then runs $\pi'$, ignoring the observation obtained from running $\pi$.
\end{definition}

\subsubsection{Pool-based policy} When analyzing the performance of our stream-based policy, we compare our policy against the optimal \emph{pool-based} policy which is allowed to select items in any order.  Note that any stream-based policy can be viewed as a special case of pool-based policy, hence, an optimal pool-based policy can not perform worse than any optimal stream-based policy.  By abuse of notation, we still use $\pi$ to represent a pool-based policy. Formally, a pool-based policy can be encoded as a partial mapping $\pi$ that maps partial realizations $\psi$ to some distribution of $E$: $\pi: 2^{E}\times O^{E} \rightarrow \mathcal{P}'(E)$. Intuitively, $\pi(\psi)$ specifies which item to select next based on the current observation $\psi$. Letting $E(\pi, \phi)$ denote the subset of items selected by a pool-based policy $\pi$ conditioned on a realization $\phi$,  the expected  utility $f_{avg}(\pi)$ of a pool-based policy $\pi$ can be written as
%
%
  \begin{eqnarray}
f_{avg}(\pi)=\mathbb{E}_{\Phi\sim p, \Pi}[f(E(\pi, \Phi), \Phi)]~\nonumber
\end{eqnarray}
where the expectation is taken over all possible realizations $\Phi$ and the internal randomness of the policy $\pi$.
Note that if $\pi$ is a pool-based policy, then for any sequence of arrivals $\sigma$,  $f_{avg}(\pi) = \mathbb{E}[f_{avg}(\pi)\mid \sigma]$. This is because the output of a pool-based policy does not depend on the sequence of arrivals.
\subsection{Problem Formulation and Additional Notations}
Our objective is to find an stream-based policy that maximizes the worst-case expected utility subject to a budget constraint $B$, i.e.,
\begin{eqnarray}
\max_{\pi\in \Omega^s}\min_{\sigma}  \mathbb{E}[f_{avg}(\pi)\mid \sigma]
\end{eqnarray}
where $\Omega^s=\{\pi\mid \forall \phi, \sigma': c(E(\pi, \Phi, \sigma'))\leq B\}$ represents a set of all feasible stream-based policies subject to a knapsack constraint $(c, B)$. That is, a feasible policy must satisfy the budget constraint under all possible realizations and sequences of arrivals.

We next introduce some additional notations and important assumptions in order to facilitate our study.
\begin{definition}[Conditional Expected Marginal Utility of an Item]
\label{def:1}
Given a utility function $f: 2^{E\times O}\rightarrow \mathbb{R}_{\geq0}$ , the conditional expected marginal utility $\Delta(e \mid  \psi)$ of an item $e$ on top of a partial realization $\psi$ is
\begin{eqnarray}
\Delta(e \mid  \psi)=\mathbb{E}_{\Phi}[f(S \cup \{e\}, \Phi)-f(S, \Phi)\mid \Phi \sim \psi]
\end{eqnarray}
where the expectation is taken over $\Phi$ with respect to $p(\phi\mid \psi)=\Pr(\Phi=\phi \mid \Phi \sim \psi)$.
\end{definition}

\begin{definition}\cite{golovin2011adaptive}[Adaptive Submodularity and Monotonicity]
A function  $f: 2^{E\times O}\rightarrow \mathbb{R}_{\geq0}$ is adaptive submodular with respect to a prior $p(\phi)$ if for any two partial realization $\psi$ and $\psi'$ such that $\psi\subseteq \psi'$ and any item $e\in E\setminus \mathrm{dom}(\psi')$,
\begin{eqnarray}
 \Delta(e \mid \psi) \geq \Delta(e\mid \psi')
\end{eqnarray}
Moreover, if $f: 2^{E\times O}\rightarrow \mathbb{R}_{\geq0}$ is adaptive monotone with respect to a prior $p(\phi)$, then we have $ \Delta(e \mid \psi) \geq 0$ for any partial realization $\psi$ and any item $e\in E\setminus \mathrm{dom}(\psi)$.
\end{definition}

\begin{definition}[Conditional Expected Marginal Utility of a Pool-based Policy]
\label{def:policy}
Given a utility function $f: 2^{E\times O}\rightarrow \mathbb{R}_{\geq0}$, the conditional expected marginal utility $\Delta(\pi \mid \psi)$ of a pool-based policy $\pi$ on top of  partial realization $\psi$ is
\[\Delta(\pi\mid \psi)=\mathbb{E}_{\Phi, \Pi}[f( E(\pi, \Phi), \Phi)-f(\mathrm{dom}(\psi), \Phi)\mid \Phi\sim \psi]\]
where the expectation is taken over $\Phi$ with respect to $p(\phi\mid \psi)=\Pr(\Phi=\phi \mid \Phi \sim \psi)$ and the internal randomness of $\pi$.
\end{definition}

We next introduce a new class of stochastic functions.
 \begin{definition}[Semi-Policywise Submodularity]
A function  $f: 2^{E\times O}\rightarrow \mathbb{R}_{\geq0}$ is semi-policywise submodular with respect to a prior $p(\phi)$ and a knapsack constraint $(c, B)$ if for any partial realization $\psi$,
\begin{eqnarray}
\ f_{avg}(\pi^*) \geq \max_{\pi\in \Omega^p} \Delta(\pi\mid  \psi)
\end{eqnarray}
where  $\Omega^p$ denotes the set of all possible pool-based policies subject to a knapsack constraint $(c,B)$, i.e., $\Omega^p=\{\pi\mid \forall \phi, c(E(\pi, \phi))\leq B\}$, and \[\pi^*\in\argmax_{\pi\in \Omega^p} \ f_{avg}(\pi)\] represents an optimal pool-based policy subject to $(c, B)$.
\end{definition}

In the rest of this paper, we always assume that our utility function  $f: 2^{E\times O}\rightarrow \mathbb{R}_{\geq0}$ is adaptive monotone, adaptive submodular and semi-policywise submodular with respect to  a prior $p(\phi)$ and a knapsack constraint $(c, B)$. In appendix, we show that this type of function can be found in a variety of important real world applications.  All missing materials are moved to appendix.

\section{Uniform Cost}
We first study the case when all items have uniform costs, i.e., $\forall e\in E, c(e)=1$. Without loss of generality, assume $B$ is some positive integer.  To solve this problem, we extend the non-adaptive solution in \cite{badanidiyuru2014streaming} to the adaptive setting.
\subsection{Algorithm Design}
\begin{algorithm}[hptb]
\caption{Online Adaptive Policy $\pi^c$}
\label{alg:1}
\begin{algorithmic}[1]
\STATE $S=\emptyset; i=1; t=1; \psi_1=\emptyset$.
\WHILE {$i \leq n$ and $|S|< B$}
\IF {$\Delta(\sigma(i)\mid \psi_t)\geq \frac{v}{2B}$}
\STATE $S\leftarrow S\cup \{\sigma(i)\}$; $\psi_{t+1}\leftarrow \psi_{t} \cup \{(\sigma(i), \Phi(\sigma(i)))\}$; $t\leftarrow t+1$;
\ENDIF
\STATE $i=i+1$;
\ENDWHILE
\RETURN $S$
\end{algorithmic}
\end{algorithm}

 Recall that $\pi^*\in\argmax_{\pi\in \Omega^p} \ f_{avg}(\pi)$ represents an optimal pool-based policy subject to a budget constraint $B$, suppose we can estimate $f_{avg}(\pi^*)$ approximately, i.e., we know a value $v$ such that $\beta\cdot f_{avg}(\pi^*) \geq v \geq \alpha\cdot f_{avg}(\pi^*)$ for some $\alpha\in[0,1]$ and $\beta\in[1,2]$. Our policy, called Online Adaptive Policy $\pi^c$, starts with an empty set $S = \emptyset$. In each subsequent iteration $i$, after observing an arriving item $\sigma(i)$, $\pi^c$ adds $\sigma(i)$ to $S$ if the marginal value of $\sigma(i)$ on top of the current partial realization $\psi_t$ is at
least $\frac{v}{2B}$; otherwise, it skips $\sigma(i)$. This process iterates until there are no more arriving items or it reaches the cardinality constraint. A detailed description of $\pi^c$ is listed in Algorithm \ref{alg:1}.

\subsection{Performance Analysis}
We present the main result of this section in the following theorem. 
\begin{theorem}
\label{thm:!}
Assuming that we know a value $v$ such that  $\beta \cdot f_{avg}(\pi^*) \geq v \geq \alpha\cdot f_{avg}(\pi^*)$ for some $\beta\in[1,2]$ and $\alpha\in[0,1]$,  we have $\mathbb{E}[f_{avg}(\pi^c)\mid \sigma] \geq \min\{\frac{\alpha}{4}, \frac{2-\beta}{4}\} f_{avg}(\pi^*)$ for any sequence of arrivals $\sigma$.
\end{theorem}

\subsection{Offline Estimation of $f_{avg}(\pi^*)$}
Recall that the design of $\pi^c$ requires that we know a good approximation of $f_{avg}(\pi^*)$. We next explain how to obtain such an estimation.  It is well known that a simple greedy pool-based policy $\pi^g$ (which is outlined in Algorithm \ref{alg:LPP1}) provides a $(1-1/e)$ approximation for the pool-based adaptive submodular maximization problem subject to a cardinality constraint \cite{golovin2011adaptive}, i.e., $f_{avg}(\pi^g)\geq (1-1/e)f_{avg}(\pi^*)$. Hence, $f_{avg}(\pi^g)$ is a good approximation of $f_{avg}(\pi^*)$. In particular, if we set $v=f_{avg}(\pi^g)$, then we have $f_{avg}(\pi^*)\geq v \geq (1-1/e)f_{avg}(\pi^*)$. This, together with Theorem \ref{thm:!}, implies that $\pi^c$ achieves a $\frac{1-1/e}{4}$ approximation ratio against $\pi^*$. One can estimate the value of $f_{avg}(\pi^g)$ by simulating $\pi^g$ on every possible realization $\phi$ to obtain $E(\pi^g, \phi)$ and letting $f_{avg}(\pi^g)=\sum_{\phi}p(\phi) f(E(\pi^g, \phi), \phi)$. When the number of possible realizations is large, one can sample a set of realizations according to $p(\phi)$ then run the simulation. Although obtaining a good estimation of $f_{avg}(\pi^g)$ may be time consuming, this only needs to be done once in an offline manner. Thus, it does not contribute to the running time of the online implementation of $\pi^c$.

\begin{algorithm}[hptb]
\caption{Offline Adaptive Greedy Policy $\pi^{g}$}
\label{alg:LPP1}
\begin{algorithmic}[1]
\STATE $S=\emptyset; t=1; \psi_1=\emptyset$.
\WHILE {$t \leq B$}
\STATE let $e'=\argmax_{e\in E} \Delta(e\mid \psi_{t})$;
\STATE $S\leftarrow S\cup \{e'\}$; $\psi_{t+1}\leftarrow \psi_{t} \cup \{(e', \Phi(e'))\}$; $t\leftarrow t+1$;
\ENDWHILE
\RETURN $S$
\end{algorithmic}
\end{algorithm}

\section{Nonuniform Cost}
We next study the general case when items have nonuniform costs.

\subsection{Algorithm Design}
\begin{algorithm}[hptb]
\caption{Online Adaptive Policy with Nonuniform Cost $\pi^k$}
\label{alg:2}
\begin{algorithmic}[1]
\STATE $S=\emptyset; t=1; i=1; \psi_1=\emptyset$.
\WHILE {$i \leq n$}
\IF {$\frac{\Delta(\sigma(i) \mid \psi_{t})}{c(\sigma(i))}\geq \frac{v}{2B}$}
\IF {$\sum_{e\in S}c(e) + c(\sigma(i))> B$}
\STATE break;
\ELSE
\STATE $S\leftarrow S\cup \{\sigma(i)\}$; $\psi_{t+1}\leftarrow \psi_{t} \cup \{(\sigma(i), \Phi(\sigma(i)))\}$; $t\leftarrow t+1$;
\ENDIF
\ENDIF
\STATE $i=i+1$
\ENDWHILE
\RETURN $S$
\end{algorithmic}
\end{algorithm}
Suppose we can estimate $f_{avg}(\pi^*)$ approximately, i.e., we know a value $v$ such that $\beta\cdot f_{avg}(\pi^*) \geq v \geq \alpha\cdot f_{avg}(\pi^*)$ for some $\alpha\in[0,1]$ and $\beta\in[1,2]$. For each $e\in E$, let $f(e)$ denote $\mathbb{E}_{\Phi}[f(\{e\}, \Phi)]$ for short. Our policy randomly selects a solution from $\{e^*\}$ and $\pi^k$ with equal probability, where $e^*=\argmax_{e\in E} f(e)$ is the best singleton and $\pi^k$, which is called  Online Adaptive Policy with Nonuniform Cost, is a density-greedy policy. Hence, the expected utility of our policy is $(f(e^*)+ \mathbb{E}[f_{avg}(\pi^k)\mid \sigma])/2$ for any given sequence of arrivals $\sigma$. We next explain the design of $\pi^k$. $\pi^k$ starts with an empty set $S = \emptyset$. In each subsequent iteration $i$, after observing an arriving item $\sigma(i)$, it adds $\sigma(i)$ to $S$ if the marginal value per unit budget of $\sigma(i)$ on top of the current realization $\psi_t$ is at
least $\frac{v}{2B}$, i.e., $\frac{\Delta(\sigma(i) \mid \psi_{t})}{c(\sigma(i))}\geq \frac{v}{2B}$, and adding $\sigma(i)$ to $S$ does not violate the budget constraint; otherwise, if $\frac{\Delta(\sigma(i) \mid \psi_{t})}{c(\sigma(i))}< \frac{v}{2B}$, $\pi^k$ skips $\sigma(i)$. This process iterates until there are no more arriving items or it reaches the first item (excluded) that violates the budget constraint. A detailed description of $\pi^k$ is listed in Algorithm \ref{alg:2}.

\subsection{Performance Analysis}
Before presenting the main theorem, we first introduce a technical lemma.
\begin{lemma}
\label{lem:2}
Assuming that we know a value $v$ such that  $\beta\cdot f_{avg}(\pi^*) \geq v \geq \alpha\cdot f_{avg}(\pi^*)$ for some $\alpha\in[0,1]$ and $\beta\in[1,2]$, we have $\max\{f(e^*), \mathbb{E}[f_{avg}(\pi^k)\mid \sigma]\} \geq \min\{\frac{\alpha}{4}, \frac{2-\beta}{4}\}f_{avg}(\pi^*)$ for any sequence of arrivals $\sigma$.
\end{lemma}

\emph{Proof:}  We first introduce an auxiliary policy $\pi^{k+}$ that follows the same procedure of $\pi^k$ except that  $\pi^{k+}$ is allowed to add the first item that violates the budget constraint. Although $\pi^{k+}$ is not necessarily feasible, we next show that the expected utility $\mathbb{E}[f_{avg}(\pi^{k+})\mid \sigma]$ of  $\pi^{k+}$ is upper bounded by $\max\{f(e^*), \mathbb{E}[f_{avg}(\pi^k)\mid \sigma]\}$ for any sequence of arrivals $\sigma$, i.e., $\mathbb{E}[f_{avg}(\pi^{k+})\mid \sigma] \leq \max\{f(e^*), \mathbb{E}[f_{avg}(\pi^k)\mid \sigma]\}$.

\begin{proposition}
\label{lem:prep}
For any sequence of arrivals $\sigma$, \[\mathbb{E}[f_{avg}(\pi^{k+})\mid \sigma] \leq \max\{f(e^*), \mathbb{E}[f_{avg}(\pi^k)\mid \sigma]\}\]
\end{proposition}

Proposition \ref{lem:prep}, whose proof is deferred to appendix, implies that to prove this lemma, it suffices to show that $\mathbb{E}[f_{avg}(\pi^{k+})\mid \sigma]\geq \min\{\frac{\alpha}{4}, \frac{2-\beta}{4}\}f_{avg}(\pi^*)$. The rest of the analysis is devoted to proving this inequality for any fixed sequence of arrivals $\sigma$. We use $\lambda=\{\psi_1^{\lambda}, \psi_2^{\lambda}, \psi_3^{\lambda},\cdots, \psi_{z^\lambda}^{\lambda}\}$ to denote a fixed run of $\pi^{k+}$, where $\psi_{t}^\lambda$ is the partial realization of the first $t$ selected items and $z^\lambda$ is the total number of selected items under $\lambda$. Let $U=\{\lambda\mid \Pr[\lambda]>0\}$ represent all possible stories of running $\pi^{k+}$, $U^+$ represent those stories where $\pi^{k+}$ meets or violates the budget, i.e., $U^+=\{\lambda\in U\mid c(\mathrm{dom}(\psi_{z^\lambda}^{\lambda}))\geq B\}$, and $U^-$ represent those stories where $\pi^{k+}$ does not use up the budget, i.e., $U^-=\{\lambda\in U\mid c(\mathrm{dom}(\psi_{z^\lambda}^{\lambda}))< B\}$. Therefore, $U=U^+ \cup U^-$. For each $\lambda$ and $t\in [z^\lambda]$, let $e^\lambda_t$ denote the $t$-th selected item under $\lambda$. Define $\psi_0^\lambda=\emptyset$ for any $\lambda$. Using the above notations, we can represent $\mathbb{E}[f_{avg}(\pi^{k+})\mid \sigma]$ as follows:
\begin{eqnarray}
&&\mathbb{E}[f_{avg}(\pi^{k+})\mid \sigma] =\sum_{\lambda\in U}\Pr[\lambda](\sum_{t\in [z^\lambda]} \Delta(e^\lambda_t\mid \psi_{t-1}^\lambda))\\
&&=\underbrace{\sum_{\lambda\in U^+}\Pr[\lambda](\sum_{t\in [z^\lambda]} \Delta(e^\lambda_t\mid \psi_{t-1}^\lambda))}_{I} + \sum_{\lambda\in U^-}\Pr[\lambda] (\sum_{t\in [z^\lambda]} \Delta(e^\lambda_t\mid \psi_{t-1}^\lambda)) \label{eq:1b}
\end{eqnarray}
Then we consider two cases. We first consider the case when $\sum_{\lambda\in U^+}\Pr[\lambda]\geq 1/2$ and show that the value of $I$ is lower bounded by $\frac{\alpha}{4} f_{avg}(\pi^*)$. According to the definition of $U^+$, we have $\sum_{t\in[z^\lambda]}c(e^\lambda_t) \geq B$ for any $\lambda\in U^+$. Moreover, recall that for all $t\in [z^\lambda]$, $\frac{\Delta(e^\lambda_t\mid \psi_{t-1}^\lambda)}{c(e^\lambda_t)}\geq \frac{v}{2B}$ due to the design of our algorithm. Therefore, for any $\lambda\in U^+$,
\begin{eqnarray}
\sum_{t\in [z^\lambda]} \Delta(e^\lambda_t\mid \psi_{t-1}^\lambda)\geq \frac{v}{2B}\times B = \frac{v}{2}
\label{eq:2b}
\end{eqnarray}
Because we assume that $\sum_{\lambda\in U^+}\Pr[\lambda]\geq 1/2$, we have
\begin{eqnarray}
\sum_{\lambda\in U^+}\Pr[\lambda](\sum_{t\in [z^\lambda]} \Delta(e^\lambda_t\mid \psi_{t-1}^\lambda)) \geq (\sum_{\lambda\in U^+}\Pr[\lambda])\times \frac{v}{2} \geq  \frac{v}{4} \geq \frac{\alpha}{4} f_{avg}(\pi^*)
\end{eqnarray}
The first inequality is due to (\ref{eq:2b}) and the third inequality is due to the assumption that $v \geq \alpha\cdot f_{avg}(\pi^*)$. We conclude that the value of $I$ (and thus $\mathbb{E}[f_{avg}(\pi^{k+})\mid \sigma]$) is no less than $\frac{\alpha}{4} f_{avg}(\pi^*)$, i.e.,
\begin{eqnarray}
\mathbb{E}[f_{avg}(\pi^{k+})\mid \sigma]\geq \frac{\alpha}{4} f_{avg}(\pi^*)\label{eq:case1}
\end{eqnarray}

We next consider the case when $\sum_{\lambda\in U^+}\Pr[\lambda]< 1/2$. We show that 
under this case,
\begin{eqnarray}\label{eq:examine}
 \mathbb{E}[f_{avg}(\pi^{k+}) \mid \sigma]\geq \frac{2-\beta}{4} f_{avg}(\pi^*)
 \end{eqnarray}

Because $f: 2^{E\times O}\rightarrow \mathbb{R}_{\geq0}$ is adaptive monotone, we have $\mathbb{E}[f_{avg}(\pi^{k+}@\pi^*)\mid \sigma]\geq f_{avg}(\pi^*)$. To prove (\ref{eq:examine}), it suffices to show that
 \[\mathbb{E}[f_{avg}(\pi^{k+}) \mid \sigma]\geq \frac{2-\beta}{4} \mathbb{E}[f_{avg}(\pi^{k+}@\pi^*)\mid \sigma]\]
  Observe that we can represent the gap between $f_{avg}(\pi^{k+}@\pi^*)$ and $f_{avg}(\pi^*)$ conditioned on $\sigma$ as follows:
\begin{eqnarray}
&&\mathbb{E}[f_{avg}(\pi^{k+}@\pi^*) - f_{avg}(\pi^{k+}) \mid \sigma] =\sum_{\lambda\in U}\Pr[\lambda](\sum_{t\in [z^\lambda]} \Delta(\pi^*\mid \psi_{z^\lambda}^{\lambda}))\\
&&=\underbrace{\sum_{\lambda\in U^+}\Pr[\lambda](\sum_{t\in [z^\lambda]} \Delta(\pi^*\mid \psi_{z^\lambda}^{\lambda}))}_{II} + \underbrace{\sum_{\lambda\in U^-}\Pr[\lambda] (\sum_{t\in [z^\lambda]} \Delta(\pi^*\mid \psi_{z^\lambda}^{\lambda}))}_{III}
\end{eqnarray}

Because  $f: 2^{E\times O}\rightarrow \mathbb{R}_{\geq0}$ is semi-policywise submodular with respect to $p(\phi)$ and $(c, B)$, we have $ \max_{\pi\in \Omega^p} \Delta(\pi\mid \psi_{z^\lambda}^{\lambda}) \leq f_{avg}(\pi^*)$. Moreover, because $\Delta(\pi^*\mid \psi_{z^\lambda}^{\lambda}) \leq \max_{\pi\in \Omega^p} \Delta(\pi\mid \psi_{z^\lambda}^{\lambda})$, we have
\begin{eqnarray}
\Delta(\pi^*\mid \psi_{z^\lambda}^{\lambda}) \leq  f_{avg}(\pi^*)
\label{eq:3}
\end{eqnarray}
It follows that
\begin{eqnarray}
\label{eq:10b}
II= \sum_{\lambda\in U^+}\Pr[\lambda](\sum_{t\in [z^\lambda]} \Delta(\pi^*\mid \psi_{z^\lambda}^{\lambda}))\leq (\sum_{\lambda\in U^+}\Pr[\lambda]) f_{avg}(\pi^*)
\end{eqnarray}

Next, we show that $III$ is upper bounded by $(\sum_{\lambda\in U^-}\Pr[\lambda])\frac{\beta}{2} f_{avg}(\pi^*)$. For any $\psi_{z^\lambda}^{\lambda}$, we number all items $e\in E$  by decreasing ratio $\frac{\Delta(e\mid \psi_{z^\lambda}^{\lambda})}{c(e)}$, i.e., $e(1)\in \arg\max_{e\in E} \frac{\Delta(e\mid \psi_{z^\lambda}^{\lambda})}{c(e)}$. Let $l=\min\{i\in \mathbb{N}\mid \sum_{j=1}^i c(e(i))\geq B\}$. Define $D(\psi_{z^\lambda}^{\lambda})=\{e(i)\in E \mid i\in[l]\}$ as the set containing the first $l$ items. Intuitively, $D(\psi_{z^\lambda}^{\lambda})$ represents a set of \emph{best-looking} items conditional on $\psi_{z^\lambda}^{\lambda}$. Consider any $e\in D(\psi_{z^\lambda}^{\lambda})$, assuming $e$ is the $i$-th item in $D(\psi_{z^\lambda}^{\lambda})$, let
\begin{equation}
x(e, \psi_{z^\lambda}^{\lambda})= \min\{1, \frac{B-\sum_{s \in \cup_{j\in[i-1]}\{e(j)\}} c(s)}{c(e)}\}~\nonumber
\end{equation}
where $ \cup_{j\in[i-1]}\{e(j)\}$ represents the first $i-1$ items in $D(\psi_{z^\lambda}^{\lambda})$.


In analogy to Lemma 1 of \cite{gotovos2015non},
\begin{eqnarray}
\label{eq:ody}
\sum_{e\in D(\psi_{z^\lambda}^{\lambda})} x(e, \psi_{z^\lambda}^{\lambda}) \Delta(e\mid \psi_{z^\lambda}^{\lambda}) \geq  \Delta(\pi^*\mid\psi_{z^\lambda}^{\lambda})
\end{eqnarray}

Note that for every $\lambda\in U^-$, we have  $\sum_{t\in[z^\lambda]}c(e^\lambda_t) < B$, that is, $\pi^{k+}$ does not use up the budget under $\lambda$. This, together with the design of $\pi^{k+}$, indicates that for any $e\in E$, its benefit-to-cost ratio  on top of $\psi_{z^\lambda}^{\lambda}$ is less than $\frac{v}{2B}$, i.e., $\frac{\Delta(e \mid \psi_{z^\lambda}^{\lambda})}{c(e)} < \frac{v}{2B}$. Therefore,
\begin{eqnarray}
\sum_{e\in D(\psi_{z^\lambda}^{\lambda})} x(e, \psi) \Delta(e\mid \psi_{z^\lambda}^{\lambda}) \leq B\times \frac{v}{2B}=\frac{v}{2}
\label{eq:5b}
\end{eqnarray}
(\ref{eq:ody}) and (\ref{eq:5b}) imply that
\begin{eqnarray}
\Delta(\pi^*\mid \psi_{z^\lambda}^{\lambda}) \leq \frac{v}{2}
\label{eq:6b}
\end{eqnarray}

We next provide an upper bound of  $III$,
\begin{eqnarray}
III&=&\sum_{\lambda\in U^-}\Pr[\lambda] (\sum_{t\in [z^\lambda]} \Delta(\pi^*\mid \psi_{z^\lambda}^{\lambda}))\leq (\sum_{\lambda\in U^-}\Pr[\lambda]) \frac{v}{2}  \\
&\leq&  (\sum_{\lambda\in U^-}\Pr[\lambda])\frac{\beta}{2} f_{avg}(\pi^*)
\label{eq:7b}
\end{eqnarray}
 where the first inequality is due to (\ref{eq:6b}) and the second inequality is due to $v \leq \beta\cdot f_{avg}(\pi^*)$.

Now we are in position to bound the value of $\mathbb{E}[f_{avg}(\pi^{k+}@\pi^*) - f_{avg}(\pi^{k+}) \mid \sigma] $,
\begin{eqnarray}
\mathbb{E}[f_{avg}(\pi^{k+}@\pi^*) - &&f_{avg}(\pi^{k+}) \mid \sigma] = II + III \\
&\leq& (\sum_{\lambda\in U^+}\Pr[\lambda]) f_{avg}(\pi^*) + (\sum_{\lambda\in U^-}\Pr[\lambda]) \frac{\beta}{2} f_{avg}(\pi^*)\\
&\leq&  \frac{1}{2} f_{avg}(\pi^*)+ \frac{1}{2}\times\frac{\beta}{2} f_{avg}(\pi^*)\\
&=& \frac{2+\beta}{4}f_{avg}(\pi^*)\label{eq:9b}
\end{eqnarray}
The first inequality is due to (\ref{eq:10b}) and (\ref{eq:7b}). The second inequality is due to $\sum_{\lambda\in U^+}\Pr[\lambda]+\sum_{\lambda\in U^-}\Pr[\lambda]=1$ and the assumptions that $\sum_{\lambda\in U^+}\Pr[\lambda] < 1/2$ and $\beta\in[1,2]$. Because $\mathbb{E}[f_{avg}(\pi^{k+}@\pi^*) \mid \sigma] \geq \mathbb{E}[f_{avg}(\pi^*) \mid \sigma]$, which is due to $f: 2^{E\times O}\rightarrow \mathbb{R}_{\geq0}$  is adaptive monotone, we have
\begin{eqnarray}
\mathbb{E}[f_{avg}(\pi^*) - f_{avg}(\pi^{k+}) \mid \sigma] &\leq& \mathbb{E}[f_{avg}(\pi^{k+}@\pi^*) - f_{avg}(\pi^{k+}) \mid \sigma] \\
&\leq& \frac{2+\beta}{4}f_{avg}(\pi^*)
\end{eqnarray}
where the second inequality is due to (\ref{eq:9b}). This,  together with the fact that $\mathbb{E}[f_{avg}(\pi^*) \mid \sigma] = f_{avg}(\pi^*)$, i.e., the optimal pool-based policy is not dependent on the sequence of arrivals, implies  (\ref{eq:examine}).

Combining the above two cases ((\ref{eq:case1}) and (\ref{eq:examine})), we have
\begin{eqnarray}
\mathbb{E}[f_{avg}(\pi^{k+}) \mid \sigma] \geq \min\{\frac{\alpha}{4}, \frac{2-\beta}{4}\}f_{avg}(\pi^*)
\end{eqnarray}

This, together with Proposition \ref{lem:prep}, immedinately conclues this lemma. $\Box$

Recall that our final policy randomly picks a solution from $\{e^*\}$ and $\pi^k$ with equal probability, thus, its expected utility is $\frac{f(e^*)+ \mathbb{E}[f_{avg}(\pi^k)\mid \sigma]}{2}$ which is lower bounded by $\frac{\max\{f(e^*), \mathbb{E}[f_{avg}(\pi^k)\mid \sigma]\}}{2}$. This, together with Lemma \ref{lem:2}, implies the following main theorem.
\begin{theorem}
\label{thm:!}
If we randomly pick a solution from $\{e^*\}$ and $\pi^k$ with equal probability, then it achieves a $\min\{\frac{\alpha}{8}, \frac{2-\beta}{8}\}$ approximation ratio against the optimal pool-based policy $\pi^*$.
\end{theorem}

\subsection{Offline Estimation of $f_{avg}(\pi^*)$}
\begin{algorithm}[hptb]
\caption{Offline Greedy Policy with Nonuniform Cost $\pi^{gn}$}
\label{alg:LPP2}
\begin{algorithmic}[1]
\STATE $S=\emptyset; t=1; \psi_1=\emptyset$.
\WHILE {$t \leq B$}
\STATE let $e'=\argmax_{e\in E} \frac{\Delta(e\mid \psi_{t})}{c(e)}$;
\IF {$\sum_{e\in S}c(e) + c(e')> B$}
\STATE break;
\ENDIF
\STATE $S\leftarrow S\cup \{e'\}$; $\psi_{t+1}\leftarrow \psi_{t} \cup \{(e', \Phi(e'))\}$; $t\leftarrow t+1$;
\ENDWHILE
\RETURN $S$
\end{algorithmic}
\end{algorithm}
To complete the design of $\pi^k$, we next explain how to estimate the utility of the optimal pool-based policy $f_{avg}(\pi^*)$. It has been shown that the better solution between $\{e^*\}$ and a pool-based density-greedy policy $\pi^{gn}$ (Algorithm \ref{alg:LPP2}) achieves  a $(1-1/e)/2$ approximation for the pool-based adaptive submodular maximization problem subject to a knapsack constraint \cite{yuan2017adaptive}, i.e., $\max\{f_{avg}(\pi^{gn}), f(e^*)\}\geq \frac{1-1/e}{2}f_{avg}(\pi^*)$. If we set $v=\max\{f_{avg}(\pi^{gn}), f(e^*)\}$ in $\pi^k$, then we have $\alpha=(1-1/e)/2$ and $\beta=1$. This, together with Theorem \ref{thm:!}, implies that $\pi^c$ achieves a $\frac{1-1/e}{16}$ approximation ratio against $\pi^*$. One can estimate the value of $f_{avg}(\pi^{gn})$ by simulating $\pi^{gn}$ on every possible realization $\phi$ to obtain $E(\pi^{gn}, \phi)$ and letting $f_{avg}(\pi^{gn})=\sum_{\phi}p(\phi) f(E(\pi^{gn}, \phi), \phi)$. To estimate the value of $f(e^*)$, one can compute the value of $f(e)$ using $f(e)=\sum_{\phi}p(\phi) f(\{e\}, \phi)$ for all $e\in E$, then return the best result as $f(e^*)$.

\bibliographystyle{splncs04}
\bibliography{reference}
\newpage
\section{Appendix}
\subsection{Proof of Theorem \ref{thm:!}} Our proof is conducted conditional on a fixed sequence of arrivals $\sigma$. Let $\lambda=\{\psi_1^\lambda, \psi_2^\lambda, \psi_3^\lambda,\cdots, \psi_{z^\lambda}^\lambda\}$ denote a fixed run of $\pi^c$, where $\psi_{t}^\lambda$ is the partial realization of the first $t$ selected items and $z^\lambda$ is the total number of selected items under $\lambda$. For any $t\in[1, z^\lambda]$, let $e^\lambda_t$ denote the $t$-th selected item under $\lambda$, i.e., $e^\lambda_t = \mathrm{dom}(\psi_t^\lambda)\setminus\mathrm{dom}(\psi_{t-1}^\lambda)$. Suppose $\Pr[\lambda]$ is the probability that $\lambda$ occurs for any $\lambda$, we can represent the expected utility of $\pi^c$ conditioned on $\sigma$ as follows:
\begin{eqnarray}
&&\mathbb{E}[f_{avg}(\pi^c)\mid \sigma] =\sum_{\lambda\in U}\Pr[\lambda](\sum_{t\in [z^\lambda]} \Delta(e^\lambda_t\mid \psi_{t-1}^\lambda))\\
&&=\underbrace{\sum_{\lambda\in U^+}\Pr[\lambda](\sum_{t\in [z^\lambda]} \Delta(e^\lambda_t\mid \psi_{t-1}^\lambda))}_{I} + \sum_{\lambda\in U^-}\Pr[\lambda] (\sum_{t\in [z^\lambda]} \Delta(e^\lambda_t\mid \psi_{t-1}^\lambda)) \label{eq:1}
\end{eqnarray}
where $U=\{\lambda\mid \Pr[\lambda]>0\}$ represents all possible stories of running $\pi^c$, $U^+$ represents those stories where $\pi^c$ selects exactly $B$ items, i.e., $U^+=\{\lambda\in U\mid z^\lambda=B\}$, and $U^-$ represents those stories where $\pi^c$ selects fewer than $B$ items, i.e., $U^-=\{\lambda\in U\mid z^\lambda< B\}$. Therefore, $U=U^+ \cup U^-$.

We prove this lemma in two cases as follows. We first consider the case when $\sum_{\lambda\in U^+}\Pr[\lambda]\geq 1/2$ and show that the value of part $I$ of $(\ref{eq:1})$ is lower bounded by $\frac{\alpha}{4} f_{avg}(\pi^*)$. According to the definition of $U^+$, we have $z^\lambda = B$ for any $\lambda\in U^+$. Moreover, recall that for all $t\in [z^\lambda]$, $\Delta(e^\lambda_t\mid \psi_{t-1}^\lambda)\geq \frac{v}{2B}$ due to the design of our algorithm. Therefore, for any $\lambda\in U^+$,
\begin{eqnarray}
\sum_{t\in [z^\lambda]} \Delta(e^\lambda_t\mid \psi_{t-1}^\lambda)\geq \frac{v}{2B}\times B = \frac{v}{2}
\label{eq:2}
\end{eqnarray}

It follows that
\begin{eqnarray}
\sum_{\lambda\in U^+}\Pr[\lambda](\sum_{t\in [z^\lambda]} \Delta(e^\lambda_t\mid \psi_{t-1}^\lambda)) \geq (\sum_{\lambda\in U^+}\Pr[\lambda])\times \frac{v}{2} \geq  \frac{v}{4} \geq \frac{\alpha}{4} f_{avg}(\pi^*)
\end{eqnarray}
The first inequality is due to (\ref{eq:2}), the second inequality is due to the assumption that $\sum_{\lambda\in U^+}\Pr[\lambda]\geq 1/2$, and the third inequality is due to the assumption that $v \geq \alpha\cdot f_{avg}(\pi^*)$. We conclude that the value of $I$ (and thus $\mathbb{E}[f_{avg}(\pi^c)\mid \sigma]$) is no less than $\frac{\alpha}{4} f_{avg}(\pi^*)$, i.e.,
\begin{eqnarray}
\mathbb{E}[f_{avg}(\pi^c)\mid \sigma] \geq I \geq \frac{\alpha}{4} f_{avg}(\pi^*)\label{eq:c1}
\end{eqnarray}

We next consider the case when $\sum_{\lambda\in U^+}\Pr[\lambda]< 1/2$. We show that $\mathbb{E}[f_{avg}(\pi^c) \mid \sigma] \geq \frac{2-\beta}{4}f_{avg}(\pi^*)$ for any sequence of arrivals $\sigma$ under this case. Observe that
\begin{eqnarray}
&&\mathbb{E}[f_{avg}(\pi^c@\pi^*) - f_{avg}(\pi^c) \mid \sigma] =\sum_{\lambda\in U}\Pr[\lambda](\sum_{t\in [z^\lambda]} \Delta(\pi^*\mid \psi_{z^\lambda}^{\lambda}))\\
&&=\underbrace{\sum_{\lambda\in U^+}\Pr[\lambda](\sum_{t\in [z^\lambda]} \Delta(\pi^*\mid \psi_{z^\lambda}^{\lambda}))}_{II} + \underbrace{\sum_{\lambda\in U^-}\Pr[\lambda] (\sum_{t\in [z^\lambda]} \Delta(\pi^*\mid \psi_{z^\lambda}^{\lambda}))}_{III}
\end{eqnarray}

Because  $f: 2^{E\times O}\rightarrow \mathbb{R}_{\geq0}$ is semi-policywise submodular with respect to $p(\phi)$ and $(c, B)$, we have $ \max_{\pi\in \Omega^p} \Delta(\pi| \psi_{z^\lambda}^{\lambda}) \leq \max_{\pi\in \Omega^p} f_{avg}(\pi)$. Moreover, because $\Delta(\pi^*\mid \psi_{z^\lambda}^{\lambda}) \leq \max_{\pi\in \Omega^p} \Delta(\pi| \psi_{z^\lambda}^{\lambda})$ and  $\max_{\pi\in \Omega^p} f_{avg}(\pi) = f_{avg}(\pi^*)$, we have
\begin{eqnarray}
\Delta(\pi^*\mid \psi_{z^\lambda}^{\lambda}) \leq  f_{avg}(\pi^*)
\label{eq:3}
\end{eqnarray}
It follows that
\begin{eqnarray}
\label{eq:10}
II=\sum_{\lambda\in U^+}\Pr[\lambda](\sum_{t\in [z^\lambda]} \Delta(\pi^*\mid \psi_{z^\lambda}^{\lambda}))\leq (\sum_{\lambda\in U^+}\Pr[\lambda]) f_{avg}(\pi^*)
\end{eqnarray}

Next, we show that $III$ is upper bounded by $(\sum_{\lambda\in U^-}\Pr[\lambda])\frac{1}{2} f_{avg}(\pi^*)$. For any final partial realization $\psi_{z^\lambda}^{\lambda}$, let $M(\psi_{z^\lambda}^{\lambda})=\arg\max_{|R|=B}\{\sum_{e\in R}  \Delta(e \mid \psi_{z^\lambda}^{\lambda})\}$ denote the set of $B$ items having the largest marginal utility on top of $\psi_{z^\lambda}^{\lambda}$. It has been shown that if $f: 2^{E\times O}\rightarrow \mathbb{R}_{\geq0}$ is adaptive submodular, then for any $\psi_{z^\lambda}^{\lambda}$,
\begin{eqnarray}
\Delta(\pi^*\mid \psi_{z^\lambda}^{\lambda}) \leq \sum_{e\in M(\psi_{z^\lambda}^{\lambda})} \Delta(e \mid \psi_{z^\lambda}^{\lambda})
\label{eq:4}
\end{eqnarray}
Recall that for every $\lambda\in U^-$, we have $z^\lambda< B$, that is, $\pi^c$ selects fewer than $B$ items under $\lambda$. This, together with the design of $\pi^c$, indicates that for any $e\in E$, its marginal utility of $e$ on top of $\psi_{z^\lambda}^{\lambda}$ is less than $\frac{v}{2B}$, i.e., $\Delta(e \mid \psi_{z^\lambda}^{\lambda}) < \frac{v}{2B}$. Therefore,
\begin{eqnarray}
\sum_{e\in M(\psi_{z^\lambda}^{\lambda})} \Delta(e \mid \psi_{z^\lambda}^{\lambda}) \leq B\times \frac{v}{2B}=\frac{v}{2}
\label{eq:5}
\end{eqnarray}
(\ref{eq:4}) and (\ref{eq:5}) imply that
\begin{eqnarray}
\Delta(\pi^*\mid \psi_{z^\lambda}^{\lambda}) \leq \frac{v}{2}
\label{eq:6}
\end{eqnarray}

We next provide an upper bound of  $III$,
\begin{eqnarray}
\sum_{\lambda\in U^-}\Pr[\lambda] (\sum_{t\in [z^\lambda]} \Delta(\pi^*\mid \psi_{z^\lambda}^{\lambda}))\leq (\sum_{\lambda\in U^-}\Pr[\lambda]) \frac{v}{2}  \leq  (\sum_{\lambda\in U^-}\Pr[\lambda])\frac{\beta}{2} f_{avg}(\pi^*)
\label{eq:7}
\end{eqnarray}
 where the first inequality is due to (\ref{eq:6}) and the second inequality is due to $v \leq \beta f_{avg}(\pi^*)$.

Now we are in position to bound the value of $\mathbb{E}[f_{avg}(\pi^c@\pi^*) - f_{avg}(\pi^c) \mid \sigma]$.
\begin{eqnarray}
&&\mathbb{E}[f_{avg}(\pi^c@\pi^*) - f_{avg}(\pi^c) \mid \sigma] = II + III \\
&\leq& (\sum_{\lambda\in U^+}\Pr[\lambda]) f_{avg}(\pi^*) + (\sum_{\lambda\in U^-}\Pr[\lambda]) \frac{\beta}{2} f_{avg}(\pi^*)\\
&\leq&  \frac{1}{2} f_{avg}(\pi^*)+ \frac{1}{2}\times  \frac{\beta}{2} f_{avg}(\pi^*)\\
&=& \frac{2+\beta}{4}f_{avg}(\pi^*)\label{eq:9}
\end{eqnarray}
The first inequality is due to (\ref{eq:10}) and (\ref{eq:7}). The second inequality is due to $\sum_{\lambda\in U^+}\Pr[\lambda]+\sum_{\lambda\in U^-}\Pr[\lambda]=1$ and the assumptions that $\sum_{\lambda\in U^+}\Pr[\lambda] < 1/2$ and $\beta\in[1,2]$. Because $\mathbb{E}[f_{avg}(\pi^{k+}@\pi^*) \mid \sigma] \geq \mathbb{E}[f_{avg}(\pi^*) \mid \sigma]$, which is due to $f: 2^{E\times O}\rightarrow \mathbb{R}_{\geq0}$  is adaptive monotone, we have
\begin{eqnarray}
\mathbb{E}[f_{avg}(\pi^*) - f_{avg}(\pi^c) \mid \sigma] &\leq& \mathbb{E}[f_{avg}(\pi^c@\pi^*) - f_{avg}(\pi^c) \mid \sigma] \\
&\leq& \frac{2+\beta}{4}f_{avg}(\pi^*)
\end{eqnarray}
where the second inequality is due to (\ref{eq:9}). This, together with the fact that $\mathbb{E}[f_{avg}(\pi^*) \mid \sigma] = f_{avg}(\pi^*)$, i.e., the optimal pool-based policy is not dependent on the sequence of arrivals,  implies that
\begin{eqnarray}
\label{eq:c2}
\mathbb{E}[f_{avg}(\pi^c) \mid \sigma] \geq \frac{2-\beta}{4}f_{avg}(\pi^*)
\end{eqnarray}
Combining the above two cases ((\ref{eq:c1}) and (\ref{eq:c2})), we have
\begin{eqnarray}
\mathbb{E}[f_{avg}(\pi^c) \mid \sigma] \geq \min\{\frac{\alpha}{4}, \frac{2-\beta}{4}\}f_{avg}(\pi^*)
\end{eqnarray}

\subsection{Proof of Proposition \ref{lem:prep}} Let $\lambda=\{\psi_1^\lambda, \psi_2^\lambda, \psi_3^\lambda,\cdots, \psi_{z^\lambda}^\lambda\}$ denote a fixed run of $\pi^{k+}$, where $\psi_{t}^\lambda$ is the partial realization of the first $t$ selected items and $z^\lambda$ is the total number of selected items under $\lambda$.  Let $U=\{\lambda\mid \Pr[\lambda]>0\}$ represent all possible stories of running $\pi^{k+}$, $U^+$ represent those stories where $\pi^{k+}$ meets or violates the budget, i.e., $U^+=\{\lambda\in U\mid c(\mathrm{dom}(\psi_{z^\lambda}^{\lambda}))\geq B\}$, and $U^-$ represent those stories where $\pi^{k+}$ does not use up the budget, i.e., $U^-=\{\lambda\in U\mid c(\mathrm{dom}(\psi_{z^\lambda}^{\lambda}))< B\}$. Therefore, $U=U^+ \cup U^-$. Using the above notations, we can represent $\mathbb{E}[f_{avg}(\pi^{k+})\mid \sigma]$ as follows:
\begin{eqnarray}
&&\mathbb{E}[f_{avg}(\pi^{k+})\mid \sigma] =\sum_{\lambda\in U}\Pr[\lambda](\sum_{t\in [z^\lambda]} \Delta(e^\lambda_t\mid \psi_{t-1}^\lambda))\\
&&=\sum_{\lambda\in U^+}\Pr[\lambda](\sum_{t\in [z^\lambda]} \Delta(e^\lambda_t\mid \psi_{t-1}^\lambda)) + \sum_{\lambda\in U^-}\Pr[\lambda] (\sum_{t\in [z^\lambda]} \Delta(e^\lambda_t\mid \psi_{t-1}^\lambda))
\end{eqnarray}
Note that the outputs of $\pi^{k+}$ and $\pi^k$ differ in at most one item. This occurs only when $\pi^{k+}$ selects some item that violates the budget constraint.  Hence, by removing the last selected item form the output of $\pi^{k+1}$ under all $\lambda\in U^+$, we obtain a lower bound on the expected utility of $\pi^k$, using the same notations as for analyzing $\mathbb{E}[f_{avg}(\pi^{k+})\mid \sigma]$, as follows:
\begin{eqnarray}
&&\mathbb{E}[f_{avg}(\pi^{k})\mid \sigma] \geq\\
&&\sum_{\lambda\in U^+}\Pr[\lambda](\sum_{t\in [z^\lambda-1]} \Delta(e^\lambda_t\mid \psi_{t-1}^\lambda)) + \sum_{\lambda\in U^-}\Pr[\lambda] (\sum_{t\in [z^\lambda]} \Delta(e^\lambda_t\mid \psi_{t-1}^\lambda))
\end{eqnarray}

Hence,
\begin{eqnarray}
&&\mathbb{E}[f_{avg}(\pi^{k})\mid \sigma] +f(e^*)\\
&&\geq \sum_{\lambda\in U^+}\Pr[\lambda](\sum_{t\in [z^\lambda-1]} \Delta(e^\lambda_t\mid \psi_{t-1}^\lambda)) \\
&&\quad\quad\quad+ \sum_{\lambda\in U^-}\Pr[\lambda] (\sum_{t\in [z^\lambda]} \Delta(e^\lambda_t\mid \psi_{t-1}^\lambda))+f(e^*)\\
&&\geq \sum_{\lambda\in U^+}\Pr[\lambda](\sum_{t\in [z^\lambda-1]} \Delta(e^\lambda_t\mid \psi_{t-1}^\lambda)+f(e^*)) \\
&&\quad\quad\quad+ \sum_{\lambda\in U^-}\Pr[\lambda] (\sum_{t\in [z^\lambda]} \Delta(e^\lambda_t\mid \psi_{t-1}^\lambda))\\
&&\geq \sum_{\lambda\in U^+}\Pr[\lambda](\sum_{t\in [z^\lambda-1]} \Delta(e^\lambda_t\mid \psi_{t-1}^\lambda)+\Delta(e^\lambda_{z^\lambda}\mid \psi_{z^\lambda-1}^\lambda)) \\
&&\quad\quad\quad+ \sum_{\lambda\in U^-}\Pr[\lambda] (\sum_{t\in [z^\lambda]} \Delta(e^\lambda_t\mid \psi_{t-1}^\lambda))\\
&&=\sum_{\lambda\in U^+}\Pr[\lambda](\sum_{t\in [z^\lambda]} \Delta(e^\lambda_t\mid \psi_{t-1}^\lambda)) + \sum_{\lambda\in U^-}\Pr[\lambda] (\sum_{t\in [z^\lambda]} \Delta(e^\lambda_t\mid \psi_{t-1}^\lambda))\\
&&=\mathbb{E}[f_{avg}(\pi^{k+})\mid \sigma]
\end{eqnarray}
The third inequality is due to the assumption that $f: 2^{E\times O}\rightarrow \mathbb{R}_{\geq0}$ is adaptive submodular, which implies that $f(e^*)\geq \Delta(e^\lambda_{z^\lambda}\mid \psi_{z^\lambda-1}^\lambda)$ for any $\lambda\in U$.

\subsection{Applications}
\label{sec:app}
In this section, we show that both adaptive submodularity  and semi-policywise submodularity can be found in several important applications. We first present the concept of policywise submodularity which is first introduced in \cite{tang2021optimal}.

\begin{definition}\cite{tang2021optimal}
A function  $f: 2^{E\times O}\rightarrow \mathbb{R}_{\geq0}$ is policywise submodular with respect to a prior $p(\phi)$ and a knapsack constraint $(c, B)$ if for any two partial realizations $\psi$ and $\psi'$ such that $\psi'\subseteq \psi$ and $c(\mathrm{dom}(\psi)) \leq B$, and any $S\subseteq E$ such that $S\cap \mathrm{dom}(\psi) =\emptyset$, we have
$
\max_{\pi\in \Omega}\Delta(\pi\mid  \psi') \geq \max_{\pi\in \Omega} \Delta(\pi\mid \psi)$,
where $\Omega = \{\pi\mid \forall \phi: c(E(\pi, \phi))\leq B-c(\mathrm{dom}(\psi)), E(\pi, \phi)\subseteq S \}$ denotes the set of feasible policies which are restricted to selecting items only from $S$.
\end{definition}

In \cite{tang2021optimal}, it has been shown that many existing adaptive submodular functions used in various applications, including  pool-based active learning \cite{golovin2011adaptive,golovin2010near,gonen2013efficient,cuong2013active}, stochastic submodular cover \cite{adibi2020submodular} and adaptive viral marketing \cite{golovin2011adaptive}, also satisfy the policywise submodularity. Our next lemma shows that policywise submodularity implies semi-policywise submodularity. This indicates that all aforementioned applications satisfy both adaptive submodularity  and semi-policywise submodularity.
\begin{lemma}
If $f: 2^{E}\times O^E\rightarrow \mathbb{R}_{\geq0}$ is policywise submodular and adaptive monotone with respect to $p(\phi)$ and all knapsack constraints, then $f: 2^{E}\times O^E\rightarrow \mathbb{R}_{\geq0}$ is semi-policywise submodular with respect to $p(\phi)$ and any knapsack constraint $(c, B)$.
\end{lemma}
\emph{Proof:}  Consider two partial realizations $\emptyset$ and $\psi$, and any knapsack constraint $(c, B)$, let $B'=B+c(\mathrm{dom}(\psi))$ and $S=E\setminus \mathrm{dom}(\psi)$. Because $\emptyset\subseteq \psi$ and we assume $f: 2^{E}\times O^E\rightarrow \mathbb{R}_{\geq0}$  is policywise submodular with respect to $p(\phi)$ and all knapsack constraints, including $(c, B')$, we have
\begin{eqnarray}
\max_{\pi\in \Omega}\Delta(\pi\mid  \emptyset) \geq \max_{\pi\in \Omega} \Delta(\pi\mid \psi)
\label{eq:5}
\end{eqnarray}
where $\Omega = \{\pi\mid \forall \phi: c(E(\pi, \phi))\leq B, E(\pi, \phi)\subseteq S \}$. Let $\pi'=\argmax_{\pi\in \Omega^p} \Delta(\pi\mid  \psi)$ represent an optimal pool-based policy subject to a knapsack constraint $(c,B)$ on top of $\psi$. Due to the definition of $B'$, it is easy to verify that $\max_{\pi\in \Omega} \Delta(\pi\mid \psi)=\Delta(\pi'\mid \psi)$. Hence, (\ref{eq:5}) indicates that
\begin{eqnarray}
\max_{\pi\in \Omega}\Delta(\pi\mid  \emptyset) \geq \Delta(\pi'\mid \psi)
\label{eq:15}
\end{eqnarray}

Moreover, because $\pi^*$ represents the best pool-based policy subject to $(c, B)$, we have $f_{avg}(\pi^*) \geq  f_{avg}(\pi')\geq \Delta(\pi'\mid \emptyset)$ where the second inequality is due to  $f: 2^{E}\times O^E\rightarrow \mathbb{R}_{\geq0}$ is adaptive submodular. This, together with (\ref{eq:15}), implies that  $f_{avg}(\pi^*) \geq \Delta(\pi'\mid \psi)$. Hence, $f: 2^{E}\times O^E\rightarrow \mathbb{R}_{\geq0}$  is semi-policywise submodular with respect to $p(\phi)$ and $(c, B)$. $\Box$
\end{document}